\documentclass{article}
\usepackage{spconf,amsmath,graphicx, xcolor, amssymb, amsthm, tabularx, multirow, booktabs}

\usepackage{color}
\usepackage{xcolor}
\definecolor{OliveGreen}{rgb}{0,0.6,0}
\definecolor{SoftRed}{rgb}{1,0.2,0.2}
\usepackage{hyperref}
\hypersetup{
     breaklinks=true,
     colorlinks   = true,
     citecolor    = OliveGreen,
     linkcolor    = SoftRed,
     urlcolor     = blue
}
\usepackage[hyperpageref]{backref}
\usepackage{url}

\usepackage{xcolor,colortbl}
\definecolor{Gray}{gray}{0.85}
\definecolor{LightCyan}{rgb}{0.88,1,1}
\newcolumntype{a}{>{\columncolor{Gray}}c}
\newcolumntype{b}{>{\columncolor{white}}c}
\title{SYNTACTICALLY GUIDED GENERATIVE EMBEDDINGS FOR ZERO-SHOT SKELETON ACTION RECOGNITION}
\name{Pranay Gupta, Divyanshu Sharma, Ravi Kiran Sarvadevabhatla}
\address{Center for Visual Information Technology\\IIIT Hyderabad, Hyderabad 500032, INDIA. \\\texttt{ravi.kiran@iiit.ac.in}\\\url{https://github.com/skelemoa/synse-zsl}}
\begin{document}
\ninept
\maketitle
\begin{abstract}
We introduce SynSE, a novel syntactically guided generative approach for Zero-Shot Learning (ZSL). Our end-to-end approach learns progressively refined generative embedding spaces constrained within and across the involved modalities (visual, language). The inter-modal constraints are defined between action sequence embedding and embeddings of Parts of Speech (PoS) tagged words in the corresponding action description. We deploy SynSE for the task of skeleton-based action sequence recognition. Our design choices enable SynSE to generalize compositionally, i.e., recognize sequences whose action descriptions contain words not encountered during training. We also extend our approach to the more challenging Generalized Zero-Shot Learning (GZSL) problem via a confidence-based gating mechanism. We are the first to present zero-shot skeleton action recognition results on the large-scale NTU-60 and NTU-120 skeleton action datasets with multiple splits. Our results demonstrate SynSE's state of the art performance in both ZSL and GZSL settings compared to strong baselines on the NTU-60 and NTU-120 datasets. 
\end{abstract}

\begin{keywords}
ZSL, skeleton action recognition, VAE, deep learning, language and vision 
\end{keywords}

\section{Introduction}
\label{sec:intro}

Advances in human action recognition have been predominantly driven by the abundance of online RGB videos. However, with the advent of accurate depth sensing technologies (e.g. Microsoft Kinect, Intel Real Sense), action recognition from 3D human skeleton data has also gained traction. Skeleton representations can be advantageous since they are compact, robustly separate the action subject (human) from background and enable privacy-preserving action capture. 

The introduction of large scale skeleton action datasets such as NTU-60~\cite{shahroudy2016ntu} and NTU-120~\cite{liu2019ntu}, have allowed researchers to develop high-performance approaches for skeleton action recognition~\cite{cheng2020skeleton, liu2020disentangling, zhang2019view, shi2019two}. However, these approaches are resource intensive, prone to overfitting and fail to generalize on classes outside the training set. Therefore, there is a strong motivation for Zero-Shot Learning (ZSL) approaches in an attempt to readily generalize across actions outside the training set. 

In ZSL, visual representations and corresponding labels for \textit{seen} classes are assumed to be available. During test time, the model is evaluated using data from \textit{unseen} classes which are not present during training. Typically, side information (e.g. class attributes) is leveraged to transfer knowledge from the seen to unseen classes. As a popular approach, ZSL approaches employ a shared embedding strategy wherein the visual (image or video) features and semantic attribute features of the corresponding class labels are projected into a common embedding space~\cite{akata2015label, frome2013devise, akata2015evaluation, norouzi2013zero, xian2018feature,hubert2017learning, mukherjee2017deep}. Generative ZSL approaches present an alternative strategy wherein unseen samples~\cite{reed2016learning} or features from unseen samples~\cite{xian2018feature} are generated using Generative Adversarial Networks(GANs). Owing to the instability in training GANs, Variational Auto-Encoders(VAEs)~\cite{kumar2018generalized, mishra2018generative,schonfeld2019generalized} have also been used for feature generation. 
 
ZSL has been previously explored for skeleton action recognition. In the only available work~\cite{jasani2019skeleton} (arXiv), embedding based methods~\cite{frome2013devise, sung2018learning} are used to align visual feature embedding of skeleton action sequence with the text embedding of the descriptive action phrase (e.g `take off jacket', `put on glasses'). The visual features are represented by the final layer features of a  skeleton action recognition model and the action phrase embedding is typically obtained by pooling the individual embeddings of words comprising the phrase. However, this approach does not enable alignment of visual embedding with respect to the individual contributors of phrase semantics - the verb (`action') and the noun(s) (`participating entities'). This inability is a major shortcoming since it does not enable generalization, i.e., being able to map the test action sequences to a description containing novel combinations of verbs and nouns, some of which might be from training action descriptions themselves.  

To address these shortcomings, we propose an approach wherein the visual embedding is aligned based on the Parts of Speech (PoS) tags (verb, noun) of the phrasal words. Instead of directly mapping the visual and PoS-wise embeddings in a discriminative setting~\cite{wray2019fine}, we use group (per-PoS, visual) specific generative models with cross-group latent objective~\cite{schonfeld2019generalized} for improved ZSL performance (Section~\ref{sec:method}). We also extend our approach to the Generalized Zero-Shot Learning (GZSL) problem, a more challenging and realistic variant of ZSL wherein good performance is required from seen \textit{and} unseen classes. We do so by incorporating a confidence-based gating mechanism. (Section~\ref{sec:gating}). 
Our approach enables state-of-the-art performance for ZSL and GZSL compared to strong baselines on the NTU-60 and the much larger NTU-120 dataset. (Section~\ref{sec:res}).

The source code and pre-trained models can be accessed at \url{https://github.com/skelemoa/synse-zsl}.

\begin{figure*}[!htb]
\centering
\includegraphics[width=\textwidth]{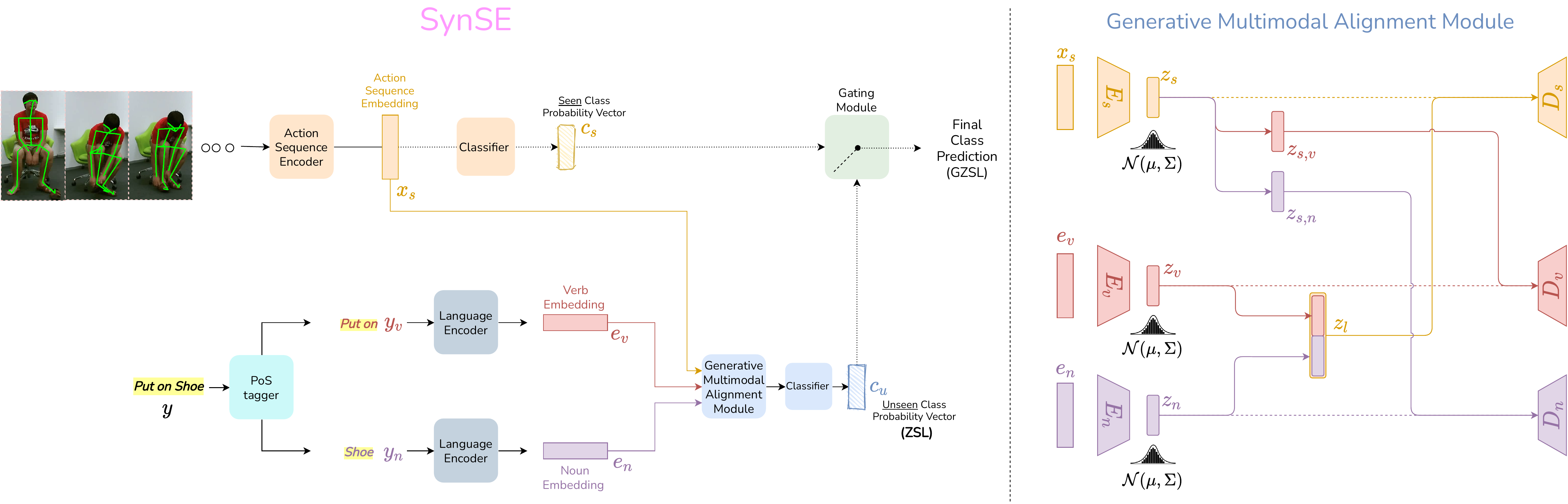}
\caption{Architectural diagram for our approach (SynSE). (left) The dotted path represents the process flow for the GZSL setting while the solid arrows represent the flow for ZSL. The Generative Multimodal Alignment Module is detailed on right side. It contains modality VAEs, where Part-of-Speech (PoS) specific latent generative embeddings $z_{v}$ (verb), $z_{n}$ (noun) are jointly aligned with segments ($z_{s,v}, z_{s,n}$) of latent generative skeleton embedding $z_{s}$ via cross-modal alignment - refer Section~\ref{sec:method} for more details. \textit{Note that the RGB images have been included for reference. Only the skeleton sequence is provided as input to the network.}}
\label{fig:alignment_module}
\end{figure*}

\section{SynSE}
\label{sec:method}

\subsection{Problem definition}

Let $ D_{tr} = \{(x_s^{tr}, y_s^{tr})\} $ denote the set of $N_{tr}$ training samples where $x_s^{tr}$ denotes visual feature embedding of a skeleton action sequence, $y_s^{tr} \in Y_{s}$ is the corresponding member from the label set of seen classes. On similar lines, $ D_{u} = \{(x_s^u, y_s^u)\} $ denotes the set of test samples with the subscript $u$ standing for unseen. Suppose $\hat{y}$ represents the test time class prediction. For ZSL, we have $ \hat{y} \in Y_{u}, Y_{s} \cap Y_{u} = \emptyset$ while for GZSL, we have $ \hat{y} \in Y_{u} \cup Y_{s}, Y_{s} \cap Y_{u} = \emptyset$. For simplicity, we drop the subscript for seen, unseen and refer to the class names as $y$ and the visual feature embedding as $x_{s}$.

\subsection{Learning modality-wise latent generative spaces}

A crucial requirement for a ZSL approach is the ability to correctly map novel inputs. For this, we employ a Variational Auto Encoder (VAE)~\cite{kingma2013auto} as the base architecture to learn the generative space of latent representations. A VAE is trained by maximizing the Evidence Lower Bound (ELBO):

\begin{equation}
    \mathcal{L} = \mathbb{E}_{q_{\phi}(z|x)}[\log p_{\theta}(x|z)] - \beta D_{KL}(q_{\phi}(z|x)||p_{\theta}(z))
\end{equation}

Here, the first term on the right hand side is the reconstruction error and the second term is the Kullback-Leibler divergence between likelihood $p_{\theta}(z)$ and the prior $q_{\phi}(z|x)$. $\beta$ is a hyperparameter which acts as a trade-off between the two error terms. A popular choice for the prior is the multivariate Gaussian distribution, $q_{\phi}(z|x) = \mathcal{N}(\mu, \Sigma)$. The VAE maps the input $x$ initially to representations for $\mu, \Sigma$ and eventually to the randomized latent representation $z$ via the reparameterization trick~\cite{kingma2013auto}.

The first stage in our approach involves learning individual latent generative latent spaces for visual and linguistic representations. This is achieved by using a VAE for each space. To enable semantically aware compositional generalization, the text description for class label $y$ is tokenized into constituent Part-of-Speech (PoS) specific sets - $y_{v}$ for verb and $y_{n}$ for noun. The tokens are encoded using a natural language encoder module to obtain the corresponding PoS-wise embeddings $e_{v}$ and $e_{n}$ (see Figure~\ref{fig:alignment_module}). Since our approach employs independent VAEs for skeleton ($s$) and linguistic ($v,n$) representations, the overall cost function for a single sample can be written as:

\begin{equation}
    \begin{aligned}
        \mathcal{L}_{VAE} = \sum_{m \in \{s, v, n\}}  \mathbb{E}_{q_{\phi}(z_{m}|x_{m})}[\log p_{\theta}(x_{m}|z_{m})] - \\
         \beta D_{KL}(q_{\phi}(z_{m}|x_{m})||p_{\theta}(z_{m}))
    \end{aligned}
\end{equation}

\subsection{Cross-modal alignment} 

The VAEs optimize latent representations for individual modalities. To achieve alignment between the skeleton sequence and linguistic latent representations, a cross-modal reconstruction objective is formulated~\cite{schonfeld2019generalized}. First, the latent embeddings from the PoS embeddings ($z_{v}$, $z_{n}$) are concatenated (see $z_{l}$ in Figure~\ref{fig:alignment_module}) and the result $x_l$ is used to reconstruct the visual representation via the skeleton representation VAE's posterior decoder $D_s$. Next, the skeleton sequence latent embedding ($z_{s}$) is uniformly mapped to as many embeddings ($z_{s,v}$, $z_{s,n}$) as the number of PoS tags. Complementary to the processing of $z_{l}$, each of the split embedding is used to reconstruct the corresponding PoS token embedding ($e_{n}$, $e_{v}$) via the corresponding PoS token embedding's decoder ($D_{v}$, $D_{n}$). Overall, the cross-modal reconstruction objective for a training sample is formulated as:

\begin{equation}
    \mathcal{L}_{CMR} = | x_{s} - D_{s}(z_{l})|_{2} + \sum_{m \in \{v, n\}} | e_{m} - D_{i}(z_{s,m})|_{2}
\end{equation}

Finally, the VAE loss and the cross-modal reconstruction loss are optimized together as: 

\begin{equation}
    \mathcal{L} = \mathcal{L}_{VAE} + \alpha \mathcal{L}_{CMR}
\end{equation}

where $\alpha$ is a trade-off weight factor.

\subsection {Zero-shot Classification using Latent Embedding} 
\label{sec:zslclassifier}

The PoS tag embeddings of each unseen class are respectively transformed by the PoS encoders ($E_{v}, E_{n}$) and used to obtain samples from the latent generative space ($z_l$ - see Figure~\ref{fig:alignment_module}). A softmax classifier $f_{u} : z_{l} \rightarrow Y_{u}$ is trained to classify these latent samples into the unseen classes.

The cross-modal VAE setup described earlier aims to align the visual features with the language features in the common latent generative space. In other words, $z_s$ and $z_l$ are optimized to be interchangeable. Taking advantage of this, during inference, the unseen class skeleton sequence representation $x_s$ is first obtained. Supplying $x_s$ to the visual VAE encoder ($E_s$) enables us to obtain the mean visual latent embedding ($\mu_{s}$) of the sequence\footnote{Note that $z_{s} = \mu_{s} + \Sigma_{s} \odot \mathcal{E}$, where $\mathcal{E} = \mathcal{N}(0 ,I)$ by the VAE reparameterization trick.}. The corresponding class prediction is obtained using $\mu_{s}$ and the classifier $f_u$ mentioned previously. 

\subsection{Gating Module for GZSL}
\label{sec:gating}

For a given skeleton sequence representation $x_s$, the probability distribution $c_s$ over seen classes is obtained from the skeleton action recognition model $f_{{s}}: x_s \rightarrow Y_{s}$ from which the action sequence embedding has been sourced all along. The unseen class classifier $f_{{u}}: E_s(x_s) \rightarrow Y_{u}$, is a part of our ZSL approach described in the previous section which provides the unseen class probabilities $c_u$. The probability distribution over \textit{all} the classes can be written as:

\begin{align}
    p(y|x) =  c_s p^{gate}(s; c_s,c_u) + c_u p^{gate}(u; c_s,c_u)
    \label{eqn:gating}
\end{align}

Further, we use a gating model (due to its superior performance for GZSL in other domains) to first decide whether the sample belongs to a seen class or an unseen class~\cite{atzmon2019adaptive}. For this, the seen and unseen class probabilities are used as features to train a probabilistic binary classifier $p^{gate}(s; c_s,c_u)$~\cite{atzmon2019adaptive}. The resulting outputs are used to determine the probability distribution over all ($Y_s \cup Y_u$) classes (Equation~\ref{eqn:gating}). 

\subsection{Implementation Details}
\label{sec:impldetails}

\begin{table}[t!]
\setlength{\tabcolsep}{5pt}
\renewcommand{\arraystretch}{1.3}
\tiny
\resizebox{\columnwidth}{!}{
\begin{tabular}{@{}ccccc@{}}
\toprule
\multirow{2}{*}{Method} & \multicolumn{2}{c}{NTU-60}  & \multicolumn{2}{c}{NTU-120} \\
  & $55/5$ split & $48/12$ split & $110/10$ split & $96/24$ split   \\
 \midrule
Jasani ~\cite{jasani2019skeleton} (preprint) & $65.53$ & - & -  & -\\
ReViSE~\cite{hubert2017learning} & $53.91$ & $17.49$ & $55.04$ & $32.38$\\
JPoSE~\cite{wray2019fine}  & $64.82$ & $28.75$ & $51.93$ & $32.44$ \\  
CADA-VAE~\cite{schonfeld2019generalized}  & $\mathbf{76.84}$ & $28.96$ & $59.53$ & $35.77$ \\ 
\midrule
SynSE (ours)  & $75.81$ & $\mathbf{33.30}$ & $\mathbf{62.69} $ & $\mathbf{38.70}$ \\
\bottomrule
\end{tabular}
}
\caption{ZSL accuracy (\%) on the NTU-60 and NTU-120 datasets.}
\label{tab:ZSL}
\end{table}

\begin{table*}[!t]
\renewcommand{\arraystretch}{1.3}
\tiny
\resizebox{\textwidth}{!}{
\begin{tabular}{@{}c bbabba|bbabba@{}}
\toprule
\multirow{2}{*}{} & \multicolumn{6}{c|}{NTU-60} & \multicolumn{6}{c}{NTU-120} \\
Method & \multicolumn{3}{c}{($55/5$) random split} & \multicolumn{3}{c|}{($48/12$) random split} &
\multicolumn{3}{c}{($110/10$) random split} &
\multicolumn{3}{c}{($96/24$) random split}  \\
\cmidrule(lr){2-4} \cmidrule(lr){5-7} \cmidrule(lr){8-10} \cmidrule(lr){11-13}
 & s & u & h & s & u & h & s & u & h & s & u & h \\ 
\midrule
ReViSE ~\cite{hubert2017learning} & $74.23$ &  $34.73$ & $29.22$ & $62.36$ & $20.77$ & $31.16$ & $ 48.69$ & $44.84$ & $46.68$ & $49.66$ &  $25.06$ & $33.31$ \\  
JPoSE ~\cite{wray2019fine}  & $64.44$ &  $50.29$ & $56.49$ & $60.49$ & $20.62$ & $30.75$ & $ 47.66 $ & $46.40$ & $47.05$ & $38.62$ & $22.79$ & $28.67$ \\  
CADA-VAE~\cite{schonfeld2019generalized} & $69.38$ & $61.79$ & $\mathbf{65.37}$ & $ 51.32$ & $27.03$ & $35.41$ & $47.16$ & $49.78$ & $48.44$ & $41.11$ & $34.14$ & $37.31$\\
\midrule
\textbf{SynSE} & $61.27$ & $56.93$ & $59.02$ & $52.21$ & $27.85$ & $\mathbf{36.33}$ & $52.51$ & $57.60$ & $\mathbf{54.94}$ & $56.39$ & $32.25$ & $\mathbf{41.04}$ \\  
SynSE (+ softgating) & $65.17$ & $59.51$ & $62.21$ & $69.23$ & $21.74$ & $33.09$ & $74.76$ & $37.68$ & $50.10$ & $72.54$ & $21.09$ & $32.67$ \\ 
SynSE (- temp. scaling) & $74.45$ & $37.46$ & $49.84$ & $45.74$ & $25.87$ & $33.05$ & $67.87$ & $38.05$ & $48.77$ & $66.97$ & $25.55$ & $36.99$ \\ 
SynSE (+ CADA-VAE's GZSL) & $82.70$ & $0$ & $0$ & $87.63$ & $0$ & $0$ & $80.43$ & $0$ & $0$ & $82.46$ & $0$ & $0$ \\ 
\bottomrule
\end{tabular}
}
\caption{GZSL Accuracy (\%) for seen (s) classes, unseen (u) classes and their harmonic mean (h) on NTU-60 and NTU-120 datasets}
\label{tab:GZSL:60}
\end{table*}

\noindent \textbf{Visual and Textual features:} The visual features $x_{s}$, are realised using the $256$ dimensional penultimate layer feature from 4s-ShiftGCN~\cite{cheng2020skeleton}, a state-of-the-art deep network for skeleton action recognition. To maintain the zero-shot assumption, we train 4s-ShiftGCN only on the seen classes. We use the Sentence BERT model~\cite{reimers2019sentence} to obtain 1024-dimensional PoS-wise word embeddings. Before splitting into verbs and nouns, the class names are modified to fill the missing PoS tag, e.g. `reading' is changed to `reading book', `drop' to `drop object', `headache' to `have headache'. For actions where adding the missing tag (usually a noun) would be unreasonable (e.g. `jump up', `stand up'), the average of all noun embeddings is used as a placeholder.

\noindent \textbf{Architectural Details:} We have a single dense layer as the encoder ($E_{s}$) and decoder ($D_{s}$), which map the input features ($x_{s}, e_{v}, e_{n}$) to the latent space ($z_{s}, z_{v}, z_{n}$) and vice versa. $x_{s}$ is $256$-dimensional and $e_{v},e_{n}$ are $1024$-dimensional. The size of the latent dimension is based on the number of unseen classes. For small number ($5$) of unseen classes, the skeleton latent dimension is set as $100$ and the latent dimension for the PoS tags is $50$. For larger number of unseen classes, the latent dimensions are doubled to $200$ and $100$ for the skeleton latent dimension and PoS tags respectively. The ZSL classifier has a single dense layer which takes latent features as input and returns the softmax probabilities for unseen classes. 

\noindent \textbf{Training Details:} The VAEs within the Generative Multimodal Alignment Module are optimized using the Adam optimizer with a learning rate of $1e^{-4}$ and a batch size of $64$. The VAEs are trained using a cyclic annealing schedule~\cite{fu2019cyclical} for multiple cycles to mitigate the vanishing KL divergence problem. The $\beta$ hyperparameter for the KL divergence is turned on after $1000$ epochs, starting with $0$ and is increased with a rate of $0.0021$ per epoch in each cycle. Similarly, the $\alpha$ parameter for the cross modal reconstruction is turned on after $1400$ epochs for experiments on NTU-60 and $1500$ epochs on NTU-120 and is kept constant with a value of $1$. One cycle is completed in $1700$ epochs for NTU-60 and $1900$ epochs for NTU-120 dataset. The zero-shot classifier (Section~\ref{sec:zslclassifier}) is also optimized using Adam with a learning rate for $1e^{-3}$. $500$ features per unseen class are generated and the classifier is trained for $300$ epochs.

The input to the gating model (Section~\ref{sec:gating}) is the concatenation of the top $k$ softmax probabilities from the outputs of the seen and unseen classifiers. We set $k$ equal to the number of unseen classes and we temperature scale~\cite{hinton2015distilling} the seen classifier probabilities as well. The gating model is implemented as a binary logistic regression classifier and optimized using LBFGS solver from the scikit-learn library with the default aggressiveness hyperparameter ($C=1$). For training the gating model, we set aside a few samples from the training set and refer to them as the gating train set. Similarly, we set aside a few samples from validation set (gating validation set). We train the gating model using the gating train set and determine the hyperparameters (temperature coefficient, threshold) using the gating validation set. The gating module is configured for use in `hard' gating mode wherein $p^{gate}(s;c_{s}, c_{u})$ and $p^{gate}(u;c_{u}, c_{u})$ in Equation~\ref{eqn:gating} take binary values~\cite{atzmon2019adaptive}.

\section{Experiments}
\label{sec:experiments}

\subsection{Datasets}

\noindent \textbf{NTU-60~\cite{shahroudy2016ntu}:} This is a large-scale dataset curated for 3D human action analysis. It contains $56{,}880$ samples belonging to $60$ action classes, with $40$ different subjects captured from $80$ distinct camera viewpoints. The action sequences of skeleton representations are in the form of 3D coordinates for $25$ human body joints. We create two splits for ZSL evaluation - a $55/5$ split with $55$ seen classes, $5$ randomly chosen unseen classes and a more challenging $48/12$ split.

\noindent \textbf{NTU-120~\cite{liu2019ntu}:} NTU-120 builds upon NTU-60 and contains $60$ additional fine-grained action classes. It contains a total of $114{,}480$ samples spread across $120$ actions performed by $106$ different subjects captured from $155$ different camera viewpoints. Analogous to the NTU-60 ZSL evaluation setup, we create two  splits - $110$ (seen)/$10$ (unseen) and $96$ (seen)/$24$ (unseen).

\subsection{Experimental Details}

We perform ZSL and GZSL experiments on the NTU-60 and NTU-120 datasets on the described splits. Since no previous works for skeleton ZSL exist, we modify representative state-of-the-art approaches from other problem domains and implement from scratch. 

CADA-VAE~\cite{schonfeld2019generalized} learns a generative latent space under a cross aligned and distribution aligned objective. Since we found the distribution alignment objective to induce instability in training, we omit it during optimization. ReViSE~\cite{hubert2017learning} aims to align the latent embeddings realised via autoencoders using a Maximum Mean Discrepancy criterion. JPose~\cite{wray2019fine} attempts to learn PoS aware embeddings of word2vec representations for video retrieval tasks. It learns a series of progressively refined embeddings under inter/intra modal constraints in a discriminative setting. For fair comparison, the visual features and PoS embeddings are the same as ones used in our approach (Section~\ref{sec:impldetails}).

\begin{table}[!t]
\renewcommand{\arraystretch}{1.3}
\centering
\resizebox{\columnwidth}{!}{%
\begin{tabular}{cccc}
\toprule
\textbf{Component} & \textbf{Default in SynSE} & \textbf{Ablation} & \textbf{Accuracy} \\ 
\midrule
Language Embedding & Sentence-BERT~\cite{reimers2019sentence} & Word2Vec~\cite{mikolov2013efficient} & $ 60.76 $ \\  
Visual Features  & 4s-ShiftGCN~\cite{cheng2020skeleton} & MS-G3D~\cite{liu2020disentangling} & $ 68.80 $ \\ 
\midrule
Latent Dimension & $100$ & $50$ & $73.83 $ \\
Latent Dimension & $100$ & $200$ & $74.67 $ \\
\midrule
Latent features & $500$ & $250$ & $73.89 $ \\
Latent features & $500$ & $1000$ & $73.82 $ \\
\midrule
  & original & & $\mathbf{75.81}$ \\
\bottomrule
\end{tabular}
}
\caption{SynSE ZSL accuracy (\%) on the NTU-60 dataset for various ablations (55/5 split).}
\label{tab:AbZSL}
\end{table}

\section{Results}
\label{sec:res}

\subsection{ZSL results}

Table~\ref{tab:ZSL} shows the ZSL results of the various approaches on the NTU-60 and NTU-120 datasets. For the $55/5$ split of NTU-60, the VAE-based generative approaches significantly outperform the discriminative embedding based approaches. SynSE's performance is comparable to that of CADA-VAE. Predictably, results on the more challenging $48/12$ split show that having a larger number of unseen classes impacts performance across the board. However, SynSE offers significant improvement over other baseline approaches, including CADA-VAE. On the larger NTU-120 dataset, SynSE outperforms other methods on both the splits. 

\subsection{GZSL results}

Since we use a gating-based strategy for GZSL in SynSE (Section~\ref{sec:gating}), we compare against other baselines by incorporating the same strategy. Specifically, the seen class classifier is kept the same while the specific baseline approach provides the corresponding unseen class probabilities. Following standard convention for GZSL, we report the average seen class accuracy (s), the average unseen class accuracy (u) and their harmonic mean (h). Table~\ref{tab:GZSL:60} shows the results for datasets and the associated pre-defined splits. Similar to the trend in ZSL for the $55/5$ split of NTU-60, SynSE performs poorer compared to CADA-VAE on the harmonic scale for the 55/5 NTU-60 split. However, it outperforms other approaches on the harmonic scale for other splits of NTU-60 and NTU-120. We also compare our hard gating strategy \cite{socher2013zero} with the soft gating based strategy \cite{atzmon2019adaptive}. The results in Table~\ref{tab:GZSL:60} show that soft gating is biased towards seen classes, resulting in poor harmonic accuracy. Additionally, Table~\ref{tab:GZSL:60} also shows the significant performance hit when temperature scaling (Sec.~\ref{sec:impldetails}) is removed~\cite{hinton2015distilling}. 

To further demonstrate the effectiveness of our GZSL strategy (i.e. gating model), we explored an alternative based on the approach used for CADA-VAE~\cite{schonfeld2019generalized}, which does not involve gating. As Table~\ref{tab:GZSL:60} shows, the resulting setup ends up too heavily skewed for seen classes and is unable to classify the unseen classes.  

\subsection{Ablations}

We perform ablation experiments on the 55/5 split of the NTU-60 dataset to analyse the importance of the building blocks of our alignment module and design choices affecting its ZSL performance. As the results show (Table~\ref{tab:AbZSL}), Sentence-BERT is a superior choice to Word2Vec~\cite{mikolov2013efficient} for embedding PoS-tagged words. Similarly, 4s-ShiftGCN provides better visual embeddings compared to another state-of-the-art skeleton action recognition model MS-G3D~\cite{liu2020disentangling}. We further ablate on the architectural choices by varying the size of the latent dimension. As shown in Table~\ref{tab:AbZSL}, we see that both an increase and decrease in the size of the latent embedding causes reduction in ZSL performance. 
In order to validate our choice of $500$ latent features per class, we experiment with varying number of latent features with results as shown in Table~\ref{tab:AbZSL}.

\section{Conclusion}
\label{sec:conclusion}

In this work, we have presented SynSE, a compositional approach for infusing latent visual representations of skeleton-based human actions with syntactic information derived from corresponding textual descriptions. We present the first set of zero-shot skeleton action recognition results on the large-scale NTU-60 and NTU-120 datasets. Our experiments show that SynSE outperforms strong baselines for ZSL and the more challenging GZSL setup. Going forward, we would like to explore the viability of SynSE for zero-shot RGB video action recognition.

\bibliographystyle{IEEE}
\bibliography{synse}

\end{document}